\pdfoutput=1
\documentclass[11pt]{article}

\usepackage[]{acl}

\usepackage{times}
\usepackage{latexsym}

\usepackage[T1]{fontenc}
\usepackage[utf8]{inputenc}
 \PassOptionsToPackage{numbers, compress,sort}{natbib}
\usepackage{hyperref}       % hyperlinks
\usepackage{url}            % simple URL typesetting
\usepackage{booktabs}       % professional-quality tables
\usepackage{amsfonts}       % blackboard math symbols
\usepackage{nicefrac}       % compact symbols for 1/2, etc.
\usepackage{microtype}      % microtypography
\usepackage{xcolor}         % colors
\usepackage[export]{adjustbox}
\usepackage{dashrule}
\usepackage{multirow}
\usepackage{graphicx}
\usepackage{amsmath}
\usepackage{amssymb}
\usepackage{CJKutf8}
\usepackage{pifont}
\usepackage[utf8]{inputenc}
\usepackage{microtype}
\usepackage{caption}
\def\ModelName{\texttt{MathOctopus}}
\def\ModelNameP{\texttt{MathOctopus}$^\mathcal{P}$}
\def\ModelNameC{\texttt{MathOctopus}$^\mathcal{C}$}
\usepackage{inconsolata}

\title{Breaking Language Barriers in Multilingual Mathematical Reasoning: Insights and Observations}

\author{
 Nuo Chen$^\spadesuit$
\quad
Zinan Zheng$^\spadesuit$
\quad 
Ning Wu$^\clubsuit$
\quad \\
{\bf Ming Gong$^\clubsuit$} 
{\bf \quad Dongmei Zhang$^\clubsuit$} 
{\bf \quad Jia Li$^\spadesuit$\thanks{ \; Corresponding author.}}\\
\\
  $^\spadesuit$Hong Kong University of Science and Technology (Guangzhou)\\ Hong Kong University of Science and Technology\\
  $^\clubsuit$Microsoft \\
    \texttt{nchen022@connect.ust.hk},
    \texttt{jialee@ust.hk}\\}

\begin{document}
\maketitle
\begin{abstract}
Existing research predominantly focuses on developing powerful large language models (LLMs) for mathematical reasoning within monolingual languages, with few explorations in preserving efficacy in a multilingual context. 
To bridge this gap, this paper pioneers exploring and training powerful  Multilingual Math Reasoning (xMR) LLMs.
Firstly, by utilizing translation, we construct the first multilingual math reasoning instruction dataset, \texttt{MGSM8KInstruct}, encompassing ten distinct languages, thus addressing the issue of training data scarcity in xMR tasks. 
Based on the collected dataset, we propose different training strategies to build powerful xMR LLMs, named \ModelName, which notably outperform conventional open-source LLMs and exhibit superiority over ChatGPT in few-shot scenarios. 
 Notably, \ModelName-13B reaches 47.6\% accuracy which exceeds ChatGPT 46.3\% on MGSM testset.
 Beyond remarkable results, we unearth several pivotal observations and insights: (1) When extending the rejection sampling strategy to the multilingual context, it proves effective for model performances, albeit limited. (2) Employing parallel corpora for math Supervised Fine-Tuning (SFT) across multiple languages not only significantly enhances model performance multilingually and  elevates their monolingual performance. This indicates that crafting multilingual corpora can be regarded as a vital strategy for enhancing model performance in a specific language, especially in mathematical reasoning tasks. For instance, \ModelName-7B improves its counterparts that trained on English from 42.4\% to 50.8\% on the GSM8K test set. Codes are available at \url{https://github.com/microsoft/MathOctopus}.
\end{abstract}

\section{Introduction}

\begin{figure}
    \centering
    \includegraphics[width=1\linewidth]{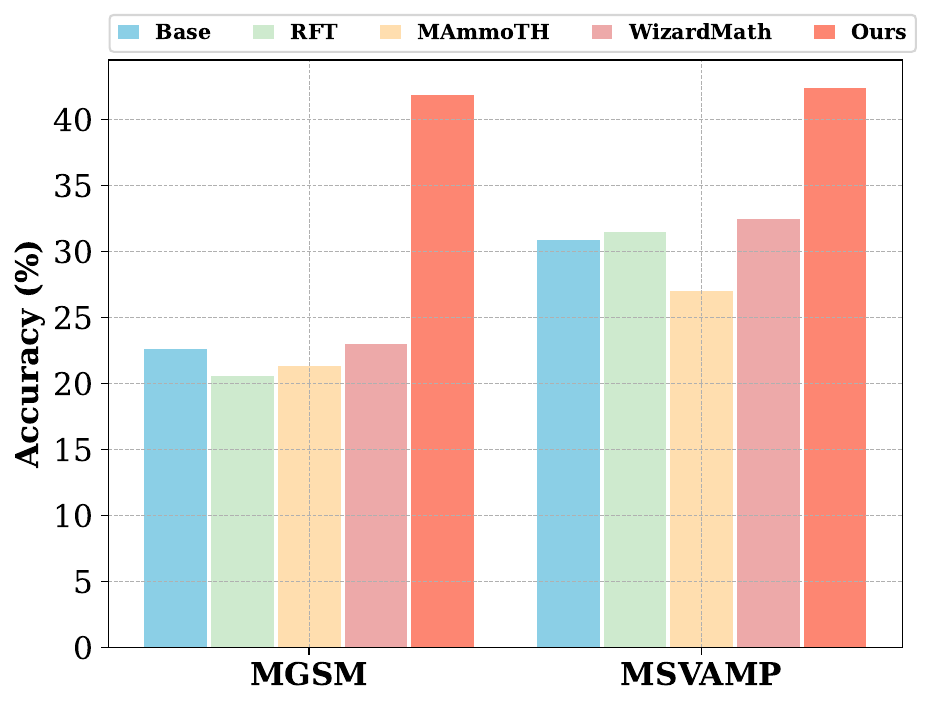}
    \caption{Different LLMs performances in MSGM and MSVAMP datasets, which all are built on LLaMA 2-7B.}
    \label{fig:motivation}
    \vspace{-15pt}
\end{figure}

Large language models (LLMs) \cite{DBLP:journals/corr/abs-2005-14165, hu2021lora, zeng2022glm, openai2023gpt4,scao2022bloom,you-etal-2022-end} such as Bloom \cite{scao2022bloom} and GPT4 \cite{openai2023gpt4} have exhibited remarkable performances across a wide array of downstream tasks. 
% They can effortlessly tackle downstream tasks by conditioning on a scant number of in-context exemplars with plain natural language task descriptions \cite{brown2020language, chen2023orca,chen2022would}. 
Notwithstanding these significant advancements, even the most extensive LLMs are confronted with challenges when faced with mathematical reasoning tasks that necessitate multiple reasoning steps \cite{gao2023pal}.

Many recent works focus on using different prompting methods like chain-of-thought (CoT) to solve mathematical problems based on close-sourced LLMs such as ChatGPT and GPT-4. Significantly,  LLaMA-Family models  \cite{touvron2023llama, touvron2023llama2} have ignited an open-source movement and rapidly reduced the disparity with these closed-source LLMs. Following this line, 
\citet{yuan2023scaling} apply rejection sampling fine-tuning (RFT) for math reasoning LLMs. WizardMath \cite{luo2023wizardmath} advances mathematical reasoning in LLMs through Reinforcement Learning from Evol-Instruct (RLEIF).
However, current efforts are primarily focusing on improving the performance of LLMs in English. Although, \citet{shi2022language} propose MGSM testset to explore multilingual math reasoning through in-context learning, training a powerful multilingual mathematical reasoning LLM  remains under-explored.

% 在XMR中，我们期待的是给出question in English or other languages，希望模型能够给出对应的答案在specific language 。这样更加方便native speaker理解，而不需要从英文再转成对应的答案进行理解。

% Their mathematical reasoning capabilities in other languages, especially low-resource ones, remain under-explored.

% Feedback (RLEIF)

\begin{figure*}[!t]
\centering
\vspace{-10pt}
\includegraphics[width=0.98\linewidth]{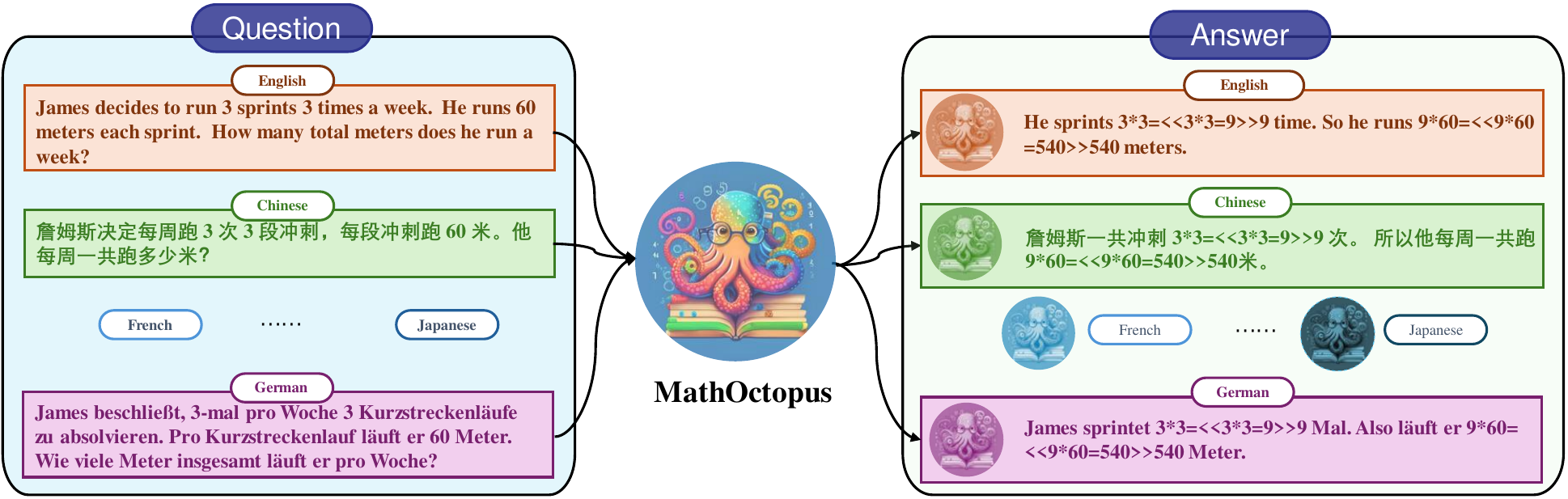}
% \vspace{-5pt}
\caption{Multilingual Math Instruction Tuning  of \ModelName.
}
\label{fig:framework}
\vspace{-10pt}
\end{figure*}

To this end, this paper empirically investigates and enhances the multilingual mathematical reasoning abilities of current open-source LLMs such as LLaMA through instruction tuning. We aim to train a single model capable of correctly answering mathematical problems in multiple target languages, not just English. 
% However, when scaling out mathematical reasoning to multiple languages, i.e., the task of
% multilingual mathematical reasoning or \textbf{xMR} for short, one main challenge is the training data scarcity in low-resource languages, where no training examples are available. 
However, the main challenge in multilingual mathematical reasoning (xMR) is the scarcity of training data in low-resource languages.
To tackle this challenge, we begin by using ChatGPT to translate the English GSM8K math training dataset into 9 various  languages. Concurrently, we employ specific rules and human verification to calibrate and align the translated corpora, ensuring the quality of data.  

The resulting data are used to construct our multilingual math instruction training dataset: \texttt{MGSM8KInstruct}, which encompasses
instructional data within two distinct settings: \textit{Parallel-training} and \textit{Cross-training}. The Parallel-training setting denotes that both the mathematical queries and the CoT answers derive from the same language. Conversely, the Cross-training setting indicates that the  questions are in English, while the corresponding answers are in other languages.
The objective of these settings is to develop 
LLMs are capable of solving mathematical problems coherently, whether presented in English or other target languages, while ensuring adaptability and maintaining rigorous mathematical reasoning across multiple languages.
One step further, to conduct a more exhaustive and comprehensive evaluation of the model’s multilingual mathematical capabilities, we additionally develop an out-of-domain xMR testset: \textbf{MSVAMP}, incorporating 10 languages, based on SVAMP \cite{patel-etal-2021-nlp}. 
% This construction involves the employment of machine translation, further refined by meticulous manual calibration.

We then use \texttt{MGSM8KInstruct} to supervise fine-tune (SFT) current open-source LLMs including LLaMA-Families ranging from 7B to 33B. Following the training, the models demonstrate exemplary abilities in xMR tasks. We name the resulting models as \ModelName, attributing their adaptability and extensive proficiency in xMR tasks across a variety of languages, as shown in Figure \ref{fig:motivation}.
% For example, \ModelName~-7B improves the average performance of LLaMA2-7B on the MGSM dataset across ten languages from 22.6$\%$ to 40$\%$. 
% after undergoing training on \texttt{MGSM8KInstruct}, the average performance of LLaMA-7B on the MGSM dataset across ten languages improves from 22.6$\%$ to 40$\%$. 
Surprisingly, compared with LLMs trained on the monolingual corpus, 
\ModelName~also shows superior performances when tested in their respective training languages.
% the specific languages compared with LLMs trained on the monolingual corpus.
For instance, 
\ModelName-7B elevate the accuracy of LLaMA2-7B on the English GSM8K from 42.4$\%$ to 50.8$\%$. We posit that this improvement is attributable to the
model's  enhanced generalization capability acquired during multilingual training, as well as the reasoning knowledge learned from other languages feeding back into in English.

% Within the realm of in-domain testing,  we notice that as the amount of data from different reasoning paths increases, the performance of the model further improves. However, xRFT can contribute fewer benefits to more performant models.

% Yet, on out-of-domain test sets, as the amount of data from different reasoning paths increases, the performance of the model may consequently decline, indicating a diminishing in their generalization and robustness.

% The resulting models de

% named \ModelName

Subsequently, we delve into an exploration of the influences of SFT data volumes and diverse reasoning paths on the efficacy of the  SFT model in multilingual context. Inspired by \cite{yuan2023scaling}, we apply multilingual rejection sampling on \ModelName~to generate different correct reasoning paths as an augmented training corpus. By incorporating this data into our prior \texttt{MGSM8KInstruct} for fine-tuning LLMs, we
observe limited enhanced multilingual mathematical reasoning outcomes. We term this training approach Multilingual Rejection Sampling Fine-tuning (xRFT). Experimentally, 
xRFT can further elevate the LLM's xMR outcomes, but it may potentially compromise the model’s generalization ability as the data amount increases.

% We further draw the following findings in in-domain (MGSM) and out-of-domain (MSVAMP) testing in terms of xRFT:
% 1) In in-domain testing, we find that as the volume of data pertaining to different reasoning paths increases, \ModelName's performance in languages other than English tends to further improve significantly, while its performance in English sees no significant change and may even potentially decline. Moreover, xRFT can contribute fewer benefits to more performant models. 2) In out-of-domain testing, with an increase in the volume of data associated with varied reasoning paths, \ModelName's xMR capabilities may consequently decline, indicating a weakening in their generalization and robustness.

In summary, our contributions are as follows:

\begin{itemize}

    \item We construct \texttt{MGSM8KInstruct}, a first multilingual math reasoning instruction dataset. Subsequently, MSVAMP, an out-of-domain multilingual mathematical reasoning dataset, is collected, serving as a robust test bed to further assess the robustness of LLMs.

    \item Based on collected data and different training strategies, we build a series of powerful LLMs, called  \ModelName~in  xMR tasks. Our model not only significantly improves its reasoning capabilities in low-resource languages compared to LLaMA but also greatly enhances its performance in English.

    \item We explore the relationship between model performance and data volume, as well as the impact of different training strategies. One of the most surprising observation is that multilingual SFT could be regarded as a crucial strategy for enhancing mathematical reasoning proficiency in LLMs.
    
    % The key in-depth analysis and observations in this paper are as follows:

    % \begin{itemize}
    % \item  Whether through parallel-training or cross-training, \ModelName~demonstrates a significant enhancement. Interestingly, the parallel-training  strategy could result in better in-domain test results while
    % the cross-training demonstrates superior generalization capabilities on out-of-domain testing.
    %     \item \ModelName~also attains significant improvements in English mathematical reasoning tasks, with cross-training yielding the optimal results. Digging deeper, we demonstrate that \ModelName~conspicuously outperforms their monolingual SFT counterparts when tested in a specific language, thereby establishing multilingual SFT as a crucial strategy for enhancing mathematical reasoning proficiencies in LLMs.
    %     \item xRFT can further elevate the LLM's xMR outcomes. However, it may potentially compromise the model’s generalization ability  as the data amount increases.
        
    %     % However, it may potentially compromise the model’s generalization ability and its performance in English as the data amount increases.
    % \end{itemize}
    
\end{itemize}

\section{Methodology}
In this section, we aim to illustrate our method in detail.
We first review the problem formulation
of multilingual mathematical reasoning. Then we describe the collection process of \texttt{MGSM8KInstruct}.
Subsequently, we present training strategies of our
 \ModelName~and multilingual rejection sampling methods, sequentially.

\subsection{Problem Formulation}
Commonly, the mathematical reasoning problem-solving task can be defined as $ \mathcal{D} = \{Q_i, O_i, A_i\}$, where $Q_i$ is the target math question, $O = \{O_1, O_2,..., O_k \}$ are answer options if $Q_i$  is a K-way multiple choice problem, $A_i$ is the corresponding ground-truth answer. Given $Q_i$  as inputs, LLMs can directly output answers or a sequence of tokens as intermediate reasoning steps $R_i$ via CoT. Then we can obtain the answer in $R_i$ through regular expression matching.

In this work, we extend mathematical reasoning tasks from monolingual to multilingual contexts: $\mathcal{D}_{en}$ to  $\mathcal{D}_{en}$...$\mathcal{D}_{zh}$. We aspire to enable only one model to successfully solve mathematical problems presented in various languages. That is, given $Q_{i(zh,es,...)}$ in a target language like Chinese, Spanish and etc, the model can furnish correct CoT responses $R_{i(zh,es,...)}$ in the specific language, even when the problems are solely presented in English. 
% The objective is to develop a model capable of understanding and solving mathematical problems across diverse linguistic mediums, emphasizing flexibility and adaptability to different languages while maintaining the integrity of mathematical reasoning.

% \subsection{Data Collection}

\subsection{MGSM8KInstruct}

\paragraph{Source Data} Prior to going further, the main concern in xMR is data scarcity of the multilingual training corpus. 
We employ GSM8K \cite{DBLP:journals/corr/abs-2110-14168}, an English dataset comprised of middle grade-school mathematical problems annotated by humans, as our fundamental data source.  According to the officially provided solutions, each problem in GSM8K necessitates a resolution process involving between two to eight steps.

\paragraph{Target Languages} As for target languages in translation, we 
choose a set of ten languages that are typologically varied from English (En), covering different language families. Similar with \cite{shi2022language}, 
 the ensemble of languages incorporated in this study comprises Bengali (Bn), Chinese (Zh), French (Fr), German (De), Japanese (Ja), Russian (Ru), Spanish (Es), Swahili (Sw),  and Thai (Th). This diverse conglomerate facilitates an exhaustive exploration into the model’s adaptability and proficiency amidst divergent linguistic architectures and typologies.

\paragraph{Translation Process}  Given the reliability and applicability of ChatGPT for translation tasks, we utilize ChatGPT to translate 7473 problems and CoT responses from the English GSM8K training set and their corresponding CoT answers into the target languages. To ensure the quality and consistency of the translations, we adopt the following strategies within the translation prompts: 
\begin{enumerate}
    \item Maintain consistent translations for names of people and places within the sentences.
    \item Preserve the mathematical formulas during translation.
    \item All numbers must be represented using Arabic numerals to facilitate cross-lingual prediction.
    \item To ensure more accurate translations, we provide two translation examples in the prompts for each language.
\end{enumerate}

\begin{table*}[t]
    \centering
    % \tiny
    \small
    \vspace{-10pt}
    % \begin{adjustbox}{width=1.0\linewidth}
    \begin{tabular}{lcccccccccc|c}
    \toprule

         % \textbf{Models} & \textbf{Swahili} & \textbf{English} & \textbf{Chinese} & \textbf{Bengali} & \textbf{German} & \textbf{Spanish} & \textbf{French} & \textbf{Japanese} & \textbf{Russian} & \textbf{Thai} & \textbf{Avg.} \\ 
    \textbf{Dataset}  &   \textbf{En} & \textbf{Sw} & \textbf{Zh} & \textbf{Bn} & \textbf{De} & \textbf{Es} & \textbf{Fr} & \textbf{Ja} & \textbf{Ru} & \textbf{Th} & \textbf{Overall} \\ 
         \midrule

\textbf{MGSM8KInstruct}&7473	&7472	&7466&	6539&	7466&	7470&	7469	&7471&	7361	&7473&	\textbf{73.6k}
 \\
\textbf{MSVAMP}&1000	&1000	&1000&1000&	1000&	1000&	1000	&1000&	1000	&1000&	\textbf{10k}
\\
               \bottomrule
    \end{tabular}
    % \end{adjustbox}
    \caption{Data statistics of our \texttt{MGSM8KInstruct} and MSVAMP.}
     \label{stastics}
     \vspace{-10pt}
\end{table*}

\paragraph{Verify Strategy} Upon inspection of our randomly sampled translations, we find that ChatGPT generally maintains semantic accuracy in translations; however, discrepancies in formula translations may arise. Thus, to uphold consistency and accuracy across multiple languages, we additionally extract all formulas present in the translated answers. If all formulas are calculated correctly and are consistent with those in English, we deem the translation to be accurate and error-free. Note that if errors persist across 5 consecutive translations, we discard that particular case.

This approach ensures a coherent and accurate translation process, allowing for comprehensive evaluation and application in xMR tasks while maintaining linguistic and mathematical integrity. 
Upon acquiring the translated data, by pairing it with the alpaca-format prompts, we are able to formulate our final training dataset \texttt{MGSM8KInstruct}, with about 73.6k samples, its statistics is shown in Table \ref{stastics}. 
Translation and training prompts are displayed in the Appendix, Table \ref{table:instruction} and \ref{table:transprompt}.

\paragraph{Crowdsourcing Verification} Although, it is common to employ machine translation to generate multilingual datasets, as seen in widely-used datasets like Xtreme \cite{hu2020xtreme}. To further verify the translation quality in our datasets, we conduct a rigorous quality check process. We randomly sample 500 samples from each  language and evaluate them for consistency (Microsoft UHRS Platform) by native speakers. The overall 91.2\% agreement rate in Table \ref{agreement} indicates reliable translation quality. 
\begin{table}[t]
    \centering
    \tiny
    % \small
   
    \begin{adjustbox}{width=1.0\linewidth}
    \begin{tabular}{l|ccccccccc}
    \toprule

         % \textbf{Models} & \textbf{Swahili} & \textbf{English} & \textbf{Chinese} & \textbf{Bengali} & \textbf{German} & \textbf{Spanish} & \textbf{French} & \textbf{Japanese} & \textbf{Russian} & \textbf{Thai} & \textbf{Avg.} \\ 
         \textbf{Lang.}  &  \textbf{Sw} & \textbf{Zh} & \textbf{Bn} & \textbf{De} & \textbf{Es} & \textbf{Fr} & \textbf{Ja} & \textbf{Ru} & \textbf{Th} \\ 
         \midrule
Agree. & 88.2 &
 90.2&
 90.6&
 90.9&
 94.9&
 94.7&
 91.3&
 90.4&
 90.3  \\

               \bottomrule
    \end{tabular}
    \end{adjustbox}
     \caption{Human agreement rate of each language.}
     \label{agreement}
     \vspace{-5pt}
\end{table}

% \subsection{MSVAMP}

% Similarly, our 

\begin{table}[t]
    \centering
    \tiny
    % \small
   
    % \begin{adjustbox}{width=1.0\linewidth}
    \begin{tabular}{l|cccccccccc}
    \toprule

         % \textbf{Models} & \textbf{Swahili} & \textbf{English} & \textbf{Chinese} & \textbf{Bengali} & \textbf{German} & \textbf{Spanish} & \textbf{French} & \textbf{Japanese} & \textbf{Russian} & \textbf{Thai} & \textbf{Avg.} \\ 
         \textbf{Times}  & \textbf{En} & \textbf{Sw} & \textbf{Zh} & \textbf{Bn} & \textbf{De} & \textbf{Es} & \textbf{Fr} & \textbf{Ja} & \textbf{Ru} & \textbf{Th} \\ 
         \midrule
10 & 1.5 &
 2.1&
 1.6&
 1.2&
 1.6&
 1.4&
 1.2&
 0.8&
 1.3&
 1.4 \\
 30 &2.5&
 3.5&
 2.7&
 1.9&
 2.6&
 2.3&
 2.0&
 1.3&
 2.2&
 2.3 \\   
50 & 3.8&
 5.2&
 4.0&
 2.9&
 3.9&
 3.5&
 3.0&
 1.9&
 3.3&
 3.4
\\
               \bottomrule
    \end{tabular}
    % \end{adjustbox}
     \caption{Distinct reasoning paths of each language with different sampling times.}
     \label{sampling}
     \vspace{-15pt}
\end{table}
\subsection{\ModelName}
\paragraph{Training Strategies}

We then use multilingual query-response pairs in \texttt{MGSM8KInstruct} to supervise fine-tune  LLMs, resulting in \ModelName, as shown in Figure \ref{fig:framework}. 
% As illustrated before,  
Let us delve into a detailed exposition of our diverse training strategies:
\begin{itemize}
    \item \textit{Parallel-training}, involves filling in the input prompts  with questions and answers in the same native language during training. This strategy is akin to teaching the model to communicate clearly in one language at a time. It helps the model get better at  answering questions accurately within the same language, making it more reliable and effective.
    \item \textit{Cross-training}, refers to our approach during training where we insert English questions and answers in one native language into the input  prompts. This approach is like mixing languages in teaching, using English questions and native language answers. It helps the model understand and connect different languages better, making it more versatile and capable of handling multilingual scenarios.
    
\end{itemize}

\begin{table*}[t]
\vspace{-10pt}
    \centering
\small
    
 \resizebox{0.95\linewidth}
{!}{
    % \begin{adjustbox}{width=1.0\linewidth}
    \begin{tabular}{l|cccccccccc|c}
    \toprule

         \textbf{Models}  & \textbf{En}  & \textbf{De} & \textbf{Es} & \textbf{Fr} &  \textbf{Sw} & \textbf{Zh} & \textbf{Bn} & \textbf{Ja} & \textbf{Ru} & \textbf{Th} & \textbf{Avg.} \\ 
         \midrule
&\multicolumn{9}{c}{Close-Source LLMs }  && \\
ChatGPT-Zero shot & 52.0 & 46.8 & 52.0 & 45.6 & 30.0  & 44.4 & 4.4 &  38.8 & 37.2 & 8.0 & 35.9
\\
ChatGPT-En 2shot &67.2 & 62.0 & 61.2& 59.2 & 40.0 & 52.8 & 7.6 &  46.8& 50.4& 15.6 & 46.3 \\
GPT4-En 2shot &\textbf{80.0} & \textbf{73.6} & \textbf{71.2} & \textbf{72.0} & \textbf{64.4} & \textbf{70.0} & \textbf{17.6}  & \textbf{71.6} & \textbf{64.0} & \textbf{40.4} & \textbf{62.5} \\
\midrule
         
       &\multicolumn{9}{c}{Open-Source LLMs (7B Model)}  && \\
       % LLaMA 2-LoRA &  27.6& 4.0&12.0& 2.0& 10.4& 18.4& 16.8& 7.6& 11.2& 3.2& 11.3 \\ 
       
        LLaMA 2 & 43.2& 37.2& 32.4& 34.4&5.2& 22.4& 3.2&  15.2& 28.0& 4.8& 22.6 \\ 
        RFT  & 44.8& 33.6& 34.0& 34.0& 2.8& 16.8& 2.4&  6.8& 29.2& 2.0& 20.6 \\ 
        MAmmoTH & 49.6& 33.2& 32.4& 32.8&2.4& 17.2& 3.6&  10.8& 26.0& 4.8& 21.3 \\ 
        
        WizardMath &  47.6& 30.4& 34.8& 30.4&3.4& 22.4& 2.0&  24.0& 30.8& 4.0& 23.0\\ 
        \midrule

\textbf{\ModelNameC} & 52.0 & 38.0 & 39.2 & 36.4 &23.6 & 31.6 & 18.8 &  27.2 & 33.6 & 21.6& 32.2 \\
\textbf{xRFT-\ModelNameC}    & 51.2 & 36.0 & 41.2 & 37.6 &24.0 & 33.2 & 18.8 &  29.6 & 36.4 & 25.2 & 33.3 \\
\midrule[0.05pt]
 % \textbf{\ModelNameP}-LoRA &  30.4& 15.2&23.6& 10.4 & 22.8 & 24.8 & 26.4 & 18.0& 22.0& 14.8 & 20.8 \\
\textbf{\ModelNameP} & 52.4& \textbf{44.8}& 42.4& \textbf{43.6}&\textbf{39.2}& 38.4& 28.8&  \textbf{36.0}& 39.6& 34.4& 40.0 \\ 
% \cmidrule

\textbf{xRFT-\ModelNameP}  & \textbf{54.8}& 43.6& \textbf{45.2}& 38.0&38.4& \textbf{45.2}& \textbf{33.2}&  35.6& \textbf{48.4}& \textbf{36.4}& \textbf{41.9} \\ 
\midrule

 &\multicolumn{9}{c}{Open-Source LLMs (13B Model)}  && \\
        LLaMA 2  & 50.4 & 42.8 & 45.2 & 40.8 &7.6 & 32.8 & 6.0 &  25.2 & 39.2 & 6.8 & 29.7 \\ 

  RFT & 52.0 & 38.4 & 46.8 & 44.8 &3.6 & 33.6 & 3.2 &  26.4 & 41.6 & 4.4 & 29.5 \\ 
        MAmmoth & \textbf{56.4}&  45.6 & 50.0 & 39.6 &  1.6 & 31.2 & 3.6 &19.2 & 36.8 & 5.2 & 28.9 \\ 
        WizardMATH & 52.8 & 40.4 & 45.6 & 42.0 &5.6 & 28.0 & 6.4 &  22.0 & 34.4 & 5.6 & 28.3 \\ 

        \midrule
\textbf{\ModelNameC}& \textbf{56.4} &47.6 & 49.6 & 47.6 & 27.2 & 39.2 & 24.0 &  40.4 & 42.0 & 24.8 & 39.9 \\
\textbf{xRFT-\ModelNameC}  & 53.6 & 48.0 & 46.4 & 46.0 &28.0 & 45.2 & 21.2 &  35.2 & 45.6 & 28.8 & 39.8 \\
\midrule
\textbf{\ModelNameP}  & 53.2 & 44.4 & 48.0 & 48.4 &42.8& 48.8 & 35.2 &  \textbf{43.2} & \textbf{47.6} & \textbf{46.8} & 45.8 \\

\textbf{xRFT-\ModelNameP} & 51.6 & \textbf{49.2} & \textbf{53.2 }& \textbf{49.6}& \textbf{46.0}& \textbf{51.2} & \textbf{42.0}  & 39.6 & \textbf{47.6} & 46.0 & \textbf{47.6} \\ 
        \midrule

 &\multicolumn{9}{c}{Open-Source LLMs (30-34B Model)}  && \\
        LLaMA 1  & 50.8 & 42.4 & 44.4 & 42.4 &3.6 & 27.6 & 3.2 &  11.6 & 38.4 & 1.2 & 26.6 \\
        RFT & \textbf{57.6} & 45.6 & 46.4 & 44.8 & 2.4 & 26.0 & 4.8 & 9.2 & 46.4 & 4.4 & 28.8 \\ 
        \midrule

       \textbf{\ModelNameC}   & 55.6 & 40.4 & 51.2 & 44.4 & 24.4 & 36.0 & 19.2 &  27.2 & 37.2 & 21.6 & 35.7  \\
            \textbf{xRFT-\ModelNameC}   & 53.6 &  47.2 & 47.6 & 44.8 &27.6 & 34.4 & 19.2 & 30.8 & 38.8 & 22.8 & 36.7 \\
            \midrule
       \textbf{\ModelNameP} & 56.4 & 47.2 & \textbf{53.2} & \textbf{48.0 }&46.8 & 52.0 & 35.2 &  39.2 & 45.6 & 41.2 & 46.5 \\ 

        \textbf{xRFT-\ModelNameP} & 51.6 & \textbf{51.2} & 52.8 & 44.4 & \textbf{47.2} & \textbf{52.4} & \textbf{37.6} &  \textbf{41.6} & \textbf{50.0} & \textbf{47.6} & \textbf{47.6}
 \\ 
% .196 & 0.104 & 0.028 & 0.088 & 0.112 & 0.104 & 0.064 & 0.1 & 0.044 & 0.0868 \\ 

        \bottomrule
    \end{tabular}
    % \end{adjustbox}
    }
    \caption{Model Performances on MGSM testset. \ModelNameP~and \ModelNameC~refer to models trained on \textit{parallel-training} and \textit{cross-training}, separately. We highlight the best results in each language of the same backbone.}
     \label{MGSM}
     \vspace{-15pt}
\end{table*}

% The first, parallel training, involves filling in the input prompts separately with questions and answers in the same native language during training.

% The second strategy, cross-training, refers to our approach during training where we insert English questions and answers in one native language into the input prompts. This methodological diversity is aimed at enhancing the model's adaptability and understanding across varied linguistic contexts.

\paragraph{Multilingual Rejection Sampling} 
Prior work \cite{yuan2023scaling} has demonstrated that LLM's performance can be further enhanced by augmenting data through rejection sampling (RFT). Consequently, in this paper, we explore whether the gains imparted by RFT persist in multilingual scenarios. After obtaining the preliminary SFT model, we perform multiple inferences with the SFT model in the \texttt{MGSM8KInstruct} dataset, sampling more diverse and accurate reasoning paths from different languages to integrate into the original dataset.
More specifically, we first eliminate samples with incorrect final answers. Subsequently, we extract all the formulas in each reasoning path and validate their accuracy; if all are correct, we consider that reasoning path as correct. We then follow the strategies from \cite{yuan2023scaling} to acquire different correct reasoning paths: a reasoning path is only collected as augmented data if no previously collected path contains identical formulas.

% It is noteworthy that different orders of elements (e.g., 9 + 8 = 7 and 8 + 9 = 7) or different orders of equations are considered different. 

However, the reasoning paths sampled from a single SFT model can be logically non-diverse. Consequently, we anticipate further enhancing the mathematical reasoning performance by leveraging reasoning paths aggregated from different models via multilingual rejection sampling. Considering the cost of prolonged inference, we currently perform 25 inferences for each language from the basic \ModelName~ 7B and 13B models respectively, meaning we sample answers for each question in every language 50 times. 
 In our experiments, we fuse all the different reasoning paths generated by the two models to obtain our final xRFT augmented data. 
 We set the temperature to 0.9 and with different seeds to expect the model to generate diverse solutions.
 The Table \ref{sampling} displays the number of different reasoning paths per question produced in each language over 50 samples. 
 % In experiments, it's observed that adding augmented data from xRFT doesn’t always improve model performance and can even reduce its generality. Detailed explanations will follow in subsequent sections.
 
% In essence, we judiciously employ multiple inferences and merge reasoning paths derived from diverse models, aiming to enrich the diversity and enhance the logical coherence of the augmented data, thereby optimizing our mathematical reasoning capabilities across different languages.

% \begin{figure}[ht]
%     \centering
%     \begin{minipage}[b]{0.45\textwidth}
%         \centering
%         \begin{tabular}{ccc}
%             \toprule
%             Header1 & Header2 & Header3 \\
%             \midrule
%             1 & 2 & 3 \\
%             4 & 5 & 6 \\
%             7 & 8 & 9 \\
%             \bottomrule
%         \end{tabular}
%         \caption{Your Table Caption}
%         \label{tab:your_table_label}
%     \end{minipage}
%     \hfill
%     \begin{minipage}[b]{0.45\textwidth}
%         \centering
%         This is an example of text that is placed side by side with a table. You can add more text, equations, or even another table or figure in this minipage.
%         \caption{figure}{Your Text Caption}
%         \label{fig:your_text_label}
%     \end{minipage}
% \end{figure}

% In essence, we scrutinize the application of RFT in a multilingual context and employ stringent validation techniques to ensure the incorporation of diverse and accurate augmented data, emphasizing the nuanced differences in formulaic representations.
\section{Experiments}

\begin{table*}[t]
\vspace{-10pt}
    \centering
    % \tiny
    \small
     \resizebox{0.95\linewidth}
{!}{
    % \begin{adjustbox}{width=1.0\linewidth}
    \begin{tabular}{l|cccccccccc|c}
    \toprule
    % \multicolumn{12}{c}{\textit{\textbf{Parallel-Test On MSVAMP}}} \\
    % \midrule
         % \textbf{Models} & \textbf{Swahili} & \textbf{English} & \textbf{Chinese} & \textbf{Bengali} & \textbf{German} & \textbf{Spanish} & \textbf{French} & \textbf{Japanese} & \textbf{Russian} & \textbf{Thai} & \textbf{Avg.} \\ 
         \textbf{Models}  & \textbf{En}  & \textbf{De} & \textbf{Es} & \textbf{Fr} &  \textbf{Sw} & \textbf{Zh} & \textbf{Bn} & \textbf{Ja} & \textbf{Ru} & \textbf{Th} & \textbf{Avg.} \\ 
         \midrule
 &\multicolumn{9}{c}{Close-Source LLMs }  && \\
ChatGPT-Zero shot & 76.1 & 66.7&
69.5 & 71.9 & 63.2 & 72.4 & 3.1 &  63.3&62.3&24.4 & 57.3 \\
ChatGPT-En 2 shot & \textbf{81.2}&73.9&74.6& 78.2 & 68.4 & 78.4 & 14.4 &74.0&70.9&46.0 &66.0\\
GPT4-En 2shot &80.1 & \textbf{78.1} & \textbf{81.5} & \textbf{83.9} & \textbf{75.7} & \textbf{78.9} & \textbf{31.2} & \textbf{74.8} & \textbf{77.9} & \textbf{68.1} & \textbf{73.0} \\
% \midrule

         \midrule
       &\multicolumn{9}{c}{Open-Source LLMs (7B ModeLs)}  && \\
       % LLaMA 2-LoRA &  27.6& 4.0&12.0& 2.0& 10.4& 18.4& 16.8& 7.6& 11.2& 3.2& 11.3 \\ 
       
        LLaMA 2 & 38.8  & 39.0 & 39.2 & 39.1 &  17.2 & 35.2 & 11.5 & 31.6 & 39.1 & 18.2 & 30.9 \\ 
        RFT  & 42.7 & 40.8 & 42.5 & 41.5 & 14.9 & 34.9 & 7.7 & 33.9 & 39.5 & 16.9 & 31.5 \\ 
        MAmmoTH & 45.1& 39.6 & 42.9 & 39.9 & 4.2  & 26.8 & 4.3 &  26.7 & 33.7 & 6.3 & 27.0 \\ 
        
        WizardMath & 48.5 & 39.2 & 44.8 & 37.7 &  10.3 & 36.3 & 16.1 & 37.9 & 37.4 & 17.0 & 32.5\\
        \midrule
\textbf{\ModelNameC} & 49.2 & \textbf{48.6} & 46.8 & 46.4& 36.6 & \textbf{43.6} & 30.2  & 42.5 & \textbf{46.7} & 34.0 & 42.5 \\ 
         \textbf{xRFT-\ModelNameC} & \textbf{49.9 }&46.5 & \textbf{47.6} &\textbf{47.3} &37.7 & 43.3 & \textbf{32.9} &  42.7 & 46.6 & 36.2&\textbf{ 43.1} \\

        \midrule
         % \textbf{\ModelNameP}-LoRA &  30.4& 15.2&23.6& 10.4 & 22.8 & 24.8 & 26.4 & 18.0& 22.0& 14.8 & 20.8 \\
        \textbf{\ModelNameP} & 46.5 & 43.5 & 45.4 & 46.0 &  40.1  & 42.5 & 29.1 & 42.5 & 45.4 & 35.7 & 41.7 \\ 
        \textbf{xRFT-\ModelNameP} & 46.8   & 43.1 & 44.5 & 45.3 & \textbf{42.3}  & 43.2 & 32.8 &  \textbf{43.2} & 42.1 & \textbf{40.5} & 42.4 \\ 
        \midrule

 &\multicolumn{9}{c}{Open-Source LLMs (13B Models)}  && \\
        LLaMA 2  & 50.9 & 46.2 & 46.1 & 47.8 & 19.8 & 43.3 & 13.9  & 41.8 & 47.8 & 23.4 & 38.1 \\ 
         RFT & 47.1 & 45.1 & 45.6 & 45.2& 19.4 & 42.3 & 12.2  & 42.4 & 46.5 & 24.8 & 37.1 \\ 
        MAmmoth & 53.4 & 52.3 & 53.9 & 53.8 & 12.9 & 47.7 & 5.0  & 42.2 & 50.7 & 13.7 & 38.6 \\ 
        WizardMATH & 56.3 & 48.7 & 50.4 & 49.4 & 12.5 & 37.0 & 13.7  & 29.5 & 43.8 & 16.3 & 35.8 \\ 
        \midrule

% EN-Only & 50.9 & 19.8 & 43.3 & 13.9 & 46.2 & 46.1 & 47.8 & 41.8 & 47.8 & 23.4 & 38.1 \\ \hline
        % V1\_Parallel & 50.7 & 43.4 & 42.6 & 31.8 & 48.4 & 49.4 & 50.6 & 41.1 & 46.9 & 39.3 & 44.4 \\ \hline
        % rft\_V3\_Parallel & 44.6 & 43.4 & 46.4 & 34.2 & 47.7 & 48.2 & 49.9 & 43.1 & 48.2 & 39.5 & 44.5
\textbf{\ModelNameC}& \textbf{56.6 }  & \textbf{50.9 }& \textbf{54.2} & \textbf{54.7}& 40.4 & 49.0 & 30.3 & \textbf{46.3 }&\textbf{ 52.4} & 35.7 & \textbf{47.1} \\
 \textbf{xRFT-\ModelNameC} & 52.9 & 50.5 & 52.8 & 51.5 & 41.9 & \textbf{49.2} & 34.1  & 45.8 & 50.2 & 35.7 & 46.5  \\ 
 \midrule
\textbf{\ModelNameP}  & 50.7& 48.4 & 49.4 & 50.6 & 43.4 & 42.6 & 31.8  & 41.1 & 46.9 & 39.3 & 44.4 \\ 
        % RFT\_V3\_Parallel & 46.0 & 51.6 & 51.2 & 42.0 & 49.2 & 53.2 & 49.6 & 39.6 & 47.6 & 46.0 & 47.6 \\ \hlin
        % \textbf{\ModelName}  & 52.0 & 44.8 & 47.6 & 42.4 & 48.0 & 52.8 & 45.6 & 38.8 & 44.8 & 40.4 & 45.7 \\ 
\textbf{xRFT-\ModelNameP} & 44.6 & 47.7 & 48.2 & 49.9 & \textbf{43.4 }& 46.4 & \textbf{34.2}  & 43.1 & 48.2 & \textbf{39.5} & 44.5 \\ 
        \midrule

 &\multicolumn{9}{c}{Open-Source LLMs (30-34B Models)}  && \\
        LLaMA 1  & 49.0 & 44.1 & 45.6 & 44.3 & 9.3 & 37.5 & 3.7  & 27.0 & 43.1 & 8.4 & 31.2 \\
        RFT & 46.8  & 46.1 & 46.8 & 46.7& 11.5 & 36.6 & 6.0 & 31.1 & 44.9 & 9.9 & 32.6  \\ 
        \midrule
\textbf{\ModelNameC} & 51.5  & \textbf{50.5} & 52.1 & \textbf{52.9} & 42.1 & 46.2 & 23.2 & 42.2 & \textbf{50.5} & 33.4 & 44.5 \\ 
 \textbf{xRFT-\ModelNameC} & 48.1& 48.7 & 50.0 & 48.9 & 42.8 & 43.6 & 23.3  & 43.4 & 44.6 & 35.5 & 42.9 \\
% \hdashline[0.5pt/5pt]
% \hdashrule[0.5ex]{\linewidth}{0.5pt}{1mm}
\midrule

\textbf{\ModelNameP} & \textbf{56.4} &  47.2 & \textbf{53.2} & 48.0& \textbf{46.8} & \textbf{52.0} & 35.2 & 39.2 & 45.6 & \textbf{41.2} & \textbf{46.5 }\\ 
        \textbf{xRFT-\ModelNameP} & 48.0  & 47.5 & 48.5 & 48.3& 42.3 & 46.1 & \textbf{36.2} & \textbf{45.8} & 47.2 & \textbf{41.2} & 45.1 
 \\ 
        \bottomrule
    \end{tabular}
    % \end{adjustbox}
    }
    \caption{Model Performances on MSVAMP testset. \ModelNameP~and \ModelNameC~refer to models trained on \textit{parallel-training} and \textit{cross-training}, separately.}
     \label{MSVAMP}
     \vspace{-10pt}
\end{table*}

In this section, we first review our in-domain evaluation dataset: MGSM, and the collection of the out-of-domain testset: MSVAMP. 
% Then we illustrate our experimental setup for training, and testing.
Subsequently, 
% we introduce multiple baselines in this work. At last, 
we present the main results and findings of our experiments.  
We illustrate our experimental setup for training and testing in Appendix \ref{setup}.
% \textbf{Of note, \ModelName$^\mathcal{C}$ and \ModelName$^\mathcal{P}$ denote our models trained on \textit{cross-training} and \textit{parallel-training}, separately.}

\subsection{Evaluation Datasets}
\paragraph{MGSM} The source data of MGSM \cite{shi2022language} is collected from a subset from GSM8K testset, and then native annotators translate the subset in English into other 10 languages. Each language branch consists of 250 test samples.

\paragraph{MSVAMP} Following \cite{yue2024mammoth}, we choose SVAMP as our out-of-domain source data. Given that the answers in the SVAMP  only contain the numerical results, we focus solely on translating the questions. To ensure high-quality translations, we use Google Translate System to convert 1,000 samples from the SVAMP into ten languages, matching the same languages in our training set. We further verify translation fidelity through crowdsourced native speaker reviews in Appendix \ref{msvamp_verify}.

\subsection{Baselines}
% \paragraph{Training and Testing} In this work,  we use open-source LLaMA-2 7B to 13B and LLaMA-1 33B as backbone models, allowing us to build \ModelName~ in multiple scales. Our codes are based on DeepSpeed and Huggingface Library. For all models, we set the learning rate, epochs and max length as 2e-5, 3 and 512. The batch sizes are set to 8, 4, 2 when models scale from 7B to 33B. 
% We keep the same prompt in Table \ref{table:instruction} for testing \ModelName. 
% Please refer to Section 3.3 for xRFT settings.
% \subsection{Baselines} 

\paragraph{Close-Source LLMs} In this paper, We consider two OpenAI's LLMs: \textbf{GPT-4} and \textbf{ChatGPT} (gpt3.5-turbo) for comparison: (1) We test \textbf{ChatGPT} with  \textit{zero-shot} prompting where none exemplars are given, but we add ``Let's think step by step.'' at the end of the inputs. (2) As \citet{shi2022language} proves including EN-CoT examples could result in better performances in xMR tasks, we additionally test them with \textit{2 shot EN-coT examples}, which are shown in Appendix Table \ref{table:testprompt}.

\paragraph{Open-Source LLMs} For fairness, we primarily compare \ModelName~with several LLaMA-based models, including \textbf{LLaMA base}, \textbf{RFT}, \textbf{MAmmoTH} and \textbf{WizardMath}. In this work, LLaMA base denotes models trained on GSM8K English corpus; RFT utilizes rejection sampling on English instruction tuning; MAmmoTH \cite{yue2024mammoth} is trained based on a variety of math instruction datasets; WizardMath \cite{luo2023wizardmath} is built on Reinforcement Learning from Evol-Instruct (RLEIF) in math reasoning. As we only consider CoT solutions in this work, we don't use the code version of these models.

\subsection{Main Results}
% \subsubsection{In-Domain Results}
Table \ref{MGSM} and Table \ref{MSVAMP} show the in-domain and out-of-domain test results of different LLMs. We run the evaluation three times and report \textit{average results}. We can draw the key  observations and insights:

\begin{table}[h]
\vspace{-10pt}
    \centering
    \small
    % \begin{minipage}{0.45\textwidth}
    %     \centering
        % Table
        
        \begin{tabular}{l|cc}
        \midrule
        \multicolumn{3}{c}{\textit{\textbf{Cross-Test}}}\\
        \midrule
        Models& MGSM8K&MSVAMP \\
           \midrule
            \ModelNameP-7B & 44.4 &47.8 \\
            \ModelNameC-7B & \textbf{47.0}&\textbf{54.2} \\
            \midrule
            \ModelNameP-13B & 47.8 &45.2 \\
            \ModelNameC-13B & \textbf{54.2}&\textbf{51.5} \\
               \midrule

            \ModelNameP-33B & 46.5 &46.6  \\
            \ModelNameC-33B & \textbf{53.3}&\textbf{49.4} \\
\bottomrule

        \end{tabular}
        \caption{Average performances of \ModelNameC~and \ModelNameP~under the \textit{Cross-Test} settings.}
        \label{cross_text}
    % \end{minipage}
    \vspace{-20pt}
\end{table}

\paragraph{LLMs struggle in xMR tasks, especially in low-resource languages.} From the tables, current open-source LLMs still suffer in xMR in terms of in-domain and out-of-domain testing. For instance,  LLMs with 7B-level only achieve about 20\%-23\% accuracy in MGSM.
Another conspicuous observation is the stark performance contrast of existing models when dealing with high-resource versus low-resource languages. This discrepancy can be largely attributed to the diminutive representation of low-resource languages in their foundational corpus. In contrast, our model adeptly rectifies this shortcoming, as evidenced by its enhanced performance in languages like Thai and Swahili.

 \paragraph{Performance Superiority  of \ModelName.} The proposed \ModelName, whether in cross-training or parallel-training, both significantly outperforms other open-source LLMs by a large margin.   For instance, when operating at the 7B-level, our model can boost the LLaMA model's efficacy from a mere 22.6\% to a commendable 41.9\% on MGSM. Furthermore, this superiority still remains as the model size escalates.
 Surprisingly, \ModelNameP-13B even surpasses ChatGPT on MGSM.
 Another interesting finding is that \textbf{ \ModelNameP~performs better in the in-domain test while \ModelNameC~shows better generalization ability in the out-of-domain test}, proving their unique advantages, separately. Delving deeper, we observe that cross-training setups could generally benefits the performances in the languages that are similar with English, like Spanish.

  \begin{table}[h]
% \vspace{-10pt}
\small
    % \hfill
    % \begin{minipage}{0.45\textwidth}
        \centering
        
        \begin{tabular}{l|cc}
        \midrule
        Models& GSM8K&SVAMP \\
           \midrule
            LLaMA 2-7B & 42.4 & 38.3 \\
            \ModelNameP-7B & 49.3 &46.8 \\
            \ModelNameC-7B & \textbf{50.8}&\textbf{49.3} \\
            \midrule
            LLaMA 2-13B & 51.0 & 50.9 \\
            \ModelNameP-13B & 55.5 &52.1 \\
            \ModelNameC-13B & \textbf{56.6}&\textbf{56.6} \\
               \midrule
            LLaMA 1-33B & 50.0 & 49.0 \\
            \ModelNameP-33B & \textbf{56.0} &\textbf{52.5}  \\
            \ModelNameC-33B & 53.7&51.5 \\
\bottomrule
        \end{tabular}
        \caption{Model performances on English datasets.}
        \label{gsm8k}
       \vspace{-10pt}
    % \end{minipage}
  \end{table}

\paragraph{Marginal Impact of xRFT.} \textbf{The xRFT's contribution to model enhancement appears to be somewhat limited.} Its effectiveness diminishes, particularly in out-of-domain test scenarios. In tests within MGSM, the xRFT's contribution to \ModelNameP~hovers around a modest 1\%-2\% average uplift. However, this figure dips below 1\% in  MSVAMP testset. More intriguingly, the augmentative effect of xRFT on \ModelNameC~appears even more subdued, with potential performance deterioration observable on the MGSM.

\vspace{-5pt}

\subsubsection{Training-Testing Consistencies}

As shown in the tables, there is a clear performance difference between models using \ModelNameC~vs. \ModelNameP, even when trained with the same amount of data in in-domain testing. A deeper look into this phenomenon revealed a mismatch between the training and testing environments. Specifically, in the MGSM and MSVAMP test sets, the data follows a parallel structure where both the question and answer are in the same language—what we call \textit{parallel-test}. This matches the parallel-training settings.

However, as presented in Table \ref{cross_text}, in our detailed experiments, when the testing environment mimics the cross-training format, called \textit{cross-test}, \ModelNameC~shows better performance compared to \ModelNameP~as model sizes increase from 7B to 33B. This highlights the critical importance of ensuring a \textbf{consistent alignment between training and testing data formats} to optimize LLMs' multilingual reasoning capabilities.

% As shown in the tables,  equally significant is the discernible performance disparity between models employing \ModelNameC~versus \ModelNameP, when subjected to the same quantum of training data  in in-domain testing.
% A deeper delve into this phenomenon revealed an incongruity between the training and testing environments. Specifically, in the MGSM and MSVAMP testsets, the data adheres to a parallel structure, where both the query and its corresponding response are couched in an identical language—a configuration we term \textit{parallel-test}. This is consistent with the parallel-training settings.

% However, in our fine-grained experiments, as presented in Table \ref{cross_text} when the testing environment mimics the cross-training format, dubbed \textit{cross-test}, \ModelNameC~demonstrate a superior performance trajectory compared to \ModelNameP~when model sizes scale from 7B to 33B.
% This underlines the pivotal importance of ensuring a \textbf{harmonious alignment between training and testing data formats }for optimizing LLMs' multilingual reasoning capability. 
% % We show detailed results of them in each language across two datasets in our Appendix.

\begin{figure}[!t]
\centering
\vspace{-10pt}
\includegraphics[width=0.98\linewidth]{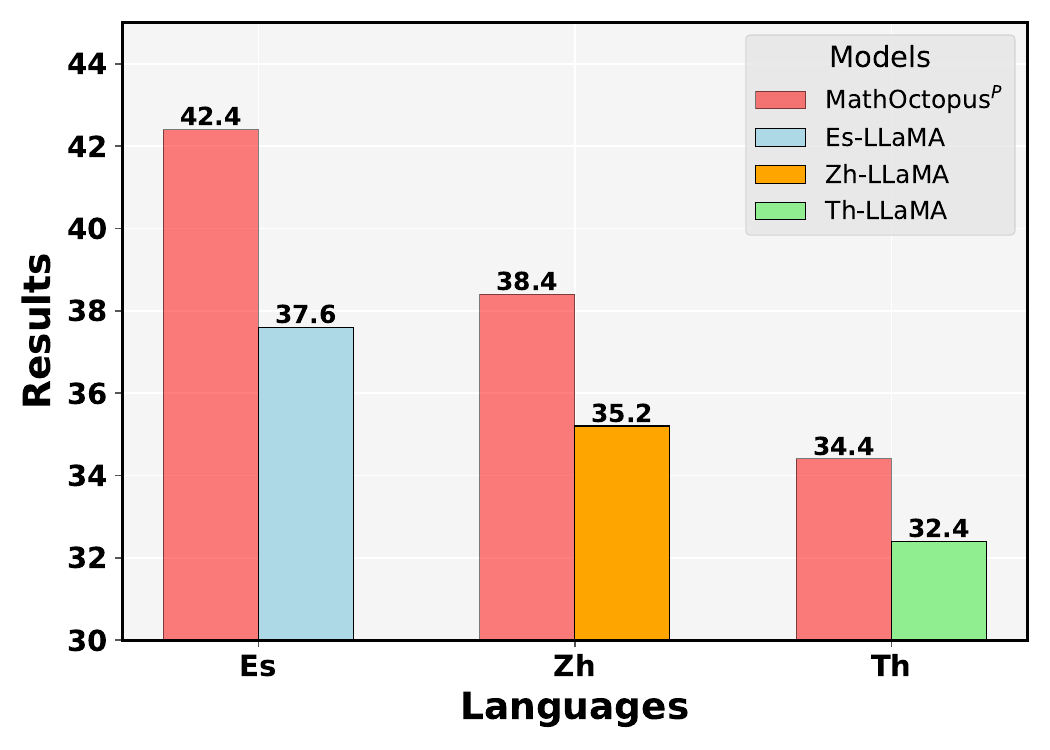}
% \vspace{-5pt}
\caption{Performances of  7B-models  on three language subsets from MGSM.
}
\label{xsft}
\vspace{-10pt}
\end{figure}

% \subsubsection{Out-of-Domain Results}

\begin{figure*}[!t]
% \vspace{-15pt}
\vspace{-10pt}
\centering
\includegraphics[width=0.98\linewidth]{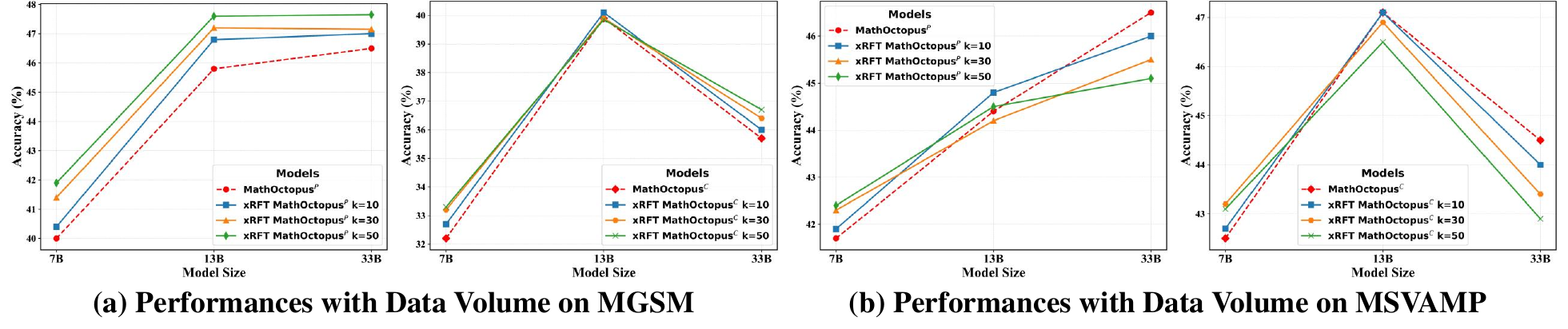}

\caption{Model performances of \ModelName~with different xRFT data volume.
}
\label{fig:XRFT}
\vspace{-15pt}
\end{figure*}

\subsubsection{ Multilingual SFT can generally benefit Monolingual SFT}
One significant observation in our experiments is that\textbf{ \ModelName~could significantly improve the performances in English.} Table \ref{gsm8k} presents the  results of LLaMA and \ModelName~on English GSM8K and SVAMP. 
Both \ModelNameP~and \ModelNameC~ substantially enhance the performance of LLaMA2 across the two datasets.  \ModelNameC~emerges as the superior performer (e.g., 50.8\% vs. 49.3\%, 49.3\% vs. 46.8\% with 7B-level). We surmise that this can be attributed to the cross-training paradigm, intensifying the model's proficiency in English comprehension.
% where during training, the model predominantly encounters queries within English  requiring reasoning in other languages. 
% This exposure, intensifies the model's proficiency in English comprehension, providing it with a distinctive edge.
% \subsubsection{Multilingual SFT can generally benefit Monolingual SFT}

% Previous experiments have demonstrated that \ModelName, when tested in English, can substantially enhance the performance of models trained in English. 
More broadly, does this situation persist in other languages as well? 
To explore this, we randomly select training sets for three languages from the training dataset: Spanish and Chinese, as well as the low-resource language Thai, and used their respective corpora to train three models, which we named Es-LLaMA, Zh-LLaMA, and Th-LLaMA, respectively. Figure \ref{xsft} separately illustrates the test results of several models in their respective training languages. We observe that our model still surpasses the results of the monolingual SFT models in their respective training languages. This suggests that, at least in the task of math reasoning, \textbf{multilingual SFT can be considered a superior training strategy to monolingual SFT, significantly elevating the model's performance in its native language.}

% \begin{minipage}[c]{0.5\textwidth}
% \includegraphics[width=\textwidth]{Images/monolingual_sft.pdf} % Include your image
% \captionof{figure}{Performances of  models-7B  on three languages subset from MGSM. }
% \label{xsft}
% \end{minipage}

% \paragraph{xRFT Sampling} We set the temperature as 0.9 and with different seeds to expect the model to generate diverse solutions.

\section{Discussion}

% Beyond the primary findings delineated above,
In this section, we  delve deeper through a series of meticulous experiments designed to address the following inquiries:
\begin{itemize}
    \item \textbf{Q1}: \textbf{The impact of xRFT}: The implications of varying xRFT data quantities?
    \item \textbf{Q2}: \textbf{Transferability of \ModelName}:
    When solely 
    utilizing corpora of specific languages from \texttt{MGSM8KInstruct} for model training, can we still observe enhancements in unseen  languages? (See Answer in Appendix \ref{discussion})
% from \texttt{MGSM8KInstruct}
    
    \item \textbf{Q3}: \textbf{Variations of the training strategy}: 
    Rather than strictly restricting the questions to English, we can propose forming pairings from the available corpus of 10 languages in \texttt{MGSM8KInstruct}, thus allowing both the questions and answers to span diverse combinations. How does the model perform under such conditions? (See Answer in Appendix \ref{discussion})
    % \item \textbf{Q4}: Can multilingual SFT  generally outperform monolingual SFT in a specific language in terms of math reasoning abilities? For instance, could \ModelName~achieve better performances than LLMs that are trained on Chinese when test in Chinese?
\end{itemize}

% Our endeavor is to shed light on these nuances and further fine-tune our understanding of the LLM's multilingual reasoning capabilities.

\subsection{RQ1: Influence of xRFT Data Volume }

In this component, we probe the impact of varying xRFT data quantities  on model performance. Figures \ref{fig:XRFT} (a) and (b) respectively illustrate the performances of models across two datasets under xRFT with sampling counts k set to \{10, 30, 50\}. 
From the visuals, it's discernible that for \ModelNameP, employing larger training corpus generally augments performance in most cases, a trend particularly pronounced in the MGSM dataset. However, these gains remain modest, especially when the backbone model becomes more performant.
In contrast, \ModelNameC~demonstrates  marginal improvements attributable to xRFT, and intriguingly, its efficacy on the MSVAMP dataset seems to wane as the k value increases. 

\textbf{This suggests that while xRFT introduces various reasoning paths, its contribution to tasks like multilingual math reasoning remains limited.} A plausible rationale is that during the multilingual SFT phase, distinct linguistic versions of the same solution might already be construed as diverse reasoning paths. Hence, multilingual SFT can be viewed as a variant of the monolingual SFT's RFT. Building upon the foundation of multilingual SFT,  supplementary benefits conferred by xRFT appear to be limited and might lead to model overfitting.

\section{Conclusion}
In this paper, we 
  pioneer the exploration of training multilingual mathematical LLMs. To address  data scarcity in low-resource languages,
we first collect the first multilingual math reasoning instruction dataset, named \texttt{MGSM8KInstruct}, consisting of ten various languages.
  The models, trained on  \texttt{MGSM8KInstruct} with different training strategies, named \ModelName, show superior performances compared to other open-source  LLMs. We prove that \ModelName~with \textit{parallel-training} could achieve better in-domain test results while \ModelName~with \textit{cross-training} presents better robustness in the collected out-of-domain test set, MSVAMP. 
  We also investigate the impact of the multilingual rejection sampling strategy, finding it has a limited effect on xMR tasks. Our extensive experiments reveal that creating aligned bilingual question-answer corpora significantly improves the model's mathematical capabilities  in its native language. In future work, we will explore additional methodologies and diverse parallel corpora for training xMR LLMs, potentially involving RLHF.
  % We investigate the influences of the multilingual rejection sampling strategy, observing it has marginal impact to xMR tasks. Based on our extensive and fine-grained experiments, we also draw several key conclusions, where the most exciting one is that creating aligned bilingual question-answer corpora can be regarded as an important way to improve model performance in its native language, especially in tasks that require mathematical reasoning. Within the picture of our future work, we will explore more methodologies and diverse parallel corpus in training xMR LLMs which may involve the RLHF and DPO.

  \section*{Limitations}
In this work, we still leave several underexplored parts, which may also contribute to building effective xMR LLMs:
\begin{itemize}
    \item Developing \ModelName~based on larger size LLMs, including LLaMA 2-70B and LLaMA-Coders, which is a future work in our following experiments.
    \item Currently, we only apply xRFT to 7B and 13B models due to the high cost of inferencing.
    We also will conduct xRFT to more performant models, further investigating its efficiency.
    \item We are still not very clear whether including more languages in \texttt{MGSM8KInstruct} could benefit current models, which will discussed in our next version.
\end{itemize}
\section*{Acknowledgement}
This work was supported by National Key Research and Development Program of China Grant No. 2023YFF0725100  and Guangzhou-HKUST(GZ) Joint Funding Scheme 2023A03J0673

% Entries for the entire Anthology, followed by custom entries
\bibliography{anthology,custom}
\clearpage
\appendix
\section{Related Works}

\paragraph{Math Reasoning with LLMs}

A pivotal metric for assessing the efficacy of LLMs is their capability in addressing intricate reasoning challenges, exemplified by mathematical reasoning tasks \cite{scao2022bloom, DBLP:journals/corr/abs-2110-14168, zhou2022least, weng2022large, chen2023teaching,chen-etal-2024-good}. Rather than yielding direct, definitive answers, prior research has illustrated that by employing a variety of prompting techniques, such as Chain-of-Thought (CoT) prompting \cite{wei2022chain}, LLMs can be guided through step-by-step reasoning, resulting in significant improvements in performance across an array of diverse reasoning tasks. \citet{DBLP:conf/acl/ImaniD023} propose the generation of multiple algebraic expressions or Python functions to solve the same mathematical problem, aiming to explore a broader spectrum of potential solutions. Additionally, \citet{li2023making} introduce a step-aware verifier to scrutinize the reasoning steps in COT, thereby enhancing the model's reasoning capabilities. Another effective approach, Self-Consistency \cite{wang2022self}, combines a wider range of solutions and derives a final answer by aggregating them to obtain the most consistent response. Meanwhile, several scholarly works have incorporated the concept of rejection sampling, in conjunction with various other techniques, to curate a more diverse set of sampled reasoning paths for the purpose of fine-tuning data augmentation \citep{huang2022large,zelikman2022star,ni2023learning,zhu-etal-2023-solving,bai2022constitutional,yuan2023rrhf,dong2023raft,llama2,song2023preference, 10.1145/3637528.3672010,you-etal-2022-end}.  
% Rejection sampling stands out as a straightforward yet highly effective fin，e-tuning augmentation method. Moreover, it finds application in aligning large language models (LLMs) with human preferences \citep{bai2022constitutional,yuan2023rrhf,dong2023raft,llama2,song2023preference}. 
Following the line, \citet{yuan2023scaling}
utilize rejection sampling to augment the data volume for fine-tuning math reasoning LLMs.

\paragraph{Instruction Tuning with LLMs}Instruction tuning serves as a pivotal component within the developmental frameworks of language models, with its primary function being to orient LLMs towards objectives that are more congruent with human preferences and functional applications \cite{chen2023large,chen2024oscarsaitheatersurvey}. The academic discourse on instruction tuning is notably concentrated on amplifying the versatile instructional capabilities of LLMs. This discourse is particularly exemplified by pioneering studies such as UnifiedQA \cite{UnifiedQA}, Zero-Prompt \cite{ZeroPrompt}, FLAN \cite{flan-t5}, and T0 \cite{T0}. These studies have embarked on an exploration into the generalization capabilities of LLMs. Following these, FLAN-v2 \cite{longpre2023flan} further investigated the impact of scaling instructional datasets on model performance. Recent innovations in this domain are veering towards employing synthetic instruction following data, distilled from models like GPT-3/4 \cite{openai2023gpt4}, to align open-source LLMs.  Recently, several works have utilized instruction tuning for training math LLMs.
\citet{yuan2023scaling,chen2022bridging} propose RFT in math reasoning, and WizardMath \cite{luo2023wizardmath} implements the "evol-instruct reinforcement learning" methodology (RLEIF), which is directed towards the refinement of prevailing math instruction data. Recently, several works \cite{Chen_MultilingualSIFT_Multilingual_Supervised_2023} extend instruction tuning from monolingual to multilingual. 
\citet{Chen_MultilingualSIFT_Multilingual_Supervised_2023} directly translate the Aplaca-GPT4 corpus to other languages and achieve great performances in MMLU tasks through multilingual instruction tuning. \citet{zhang2023bayling} further boost small MLLMs through interactive instruction tuning translation task.
However, almost all of them aim to improve mathematical reasoning in English or general multilingual generation abilities, leaving multilingual mathematical reasoning less explored. This paper aims to fill this gap by exploring effective methods for training robust LLMs in multilingual mathematical reasoning.

% In recent times, large language models (LLMs) have demonstrated remarkable abilities in handling complex reasoning tasks \cite{scao2022bloom, DBLP:journals/corr/abs-2110-14168, zhou2022least, weng2022large, chen2023teaching}. Rather than providing direct final answers as outputs, prior research has shown that by employing diverse prompting methods such as Chain-of-Thought (COT) prompting \cite{wei2022chain}, LLMs can be guided through step-by-step reasoning, resulting in notably improved performance across a wide array of reasoning tasks. \citet{DBLP:conf/acl/ImaniD023} propose to generate multiple algebraic expressions or Python functions to solve the same math problem, aiming to explore different potential solutions. \citet{li2023making}
% introduce a step-aware verifier to check the reasoning steps in COT, improving the reasoning capabilities. 
% Self-Consistency \cite{wang2022self} is another effective work that combines different solutions and gets a final answer by aggregating to retrieve the most consistent answer. 

% Among them, self-consistency bears resemblance to our proposed TIP, but our main distinction lies in the following: Given several solutions, our TIP can re-generate a new answer after cross-validating these solutions. In contrast, self-consistency can only select the most consistent answer from the existing ones.

\section{Experimental Setup}
\label{setup}
\paragraph{Training and Testing} In this work,  we use open-source LLaMA-2 7B to 13B and LLaMA-1 33B as backbone models, allowing us to build \ModelName~ in multiple scales. Our codes are based on DeepSpeed and Huggingface Library. For all models, we set the learning rate, epochs and max length as 2e-5, 3 and 512. The batch sizes are set to 8, 4, 2 when models scale from 7B to 33B.  During testing, we set the maximum output token as 512 with temperature as 0.0 to keep stable performances.
We keep the same prompt in Table \ref{table:instruction} for testing \ModelName. 
Please refer to Section 3.3 for xRFT settings.

\section{CrowdSourcing Verification of MSVAMP}
\label{msvamp_verify}
Similarly, we further verify the translation quality of MSVAMP. we sample 500 samples from each language and employ native speakers from Microsoft UHRS Platform to check the semantic consistency. We report the human agreement rates in Table \ref{nmsvamp_agreement}
\begin{table}[t]
    \centering
    \tiny
    % \small
   
    \begin{adjustbox}{width=1.0\linewidth}
    \begin{tabular}{l|ccccccccc}
    \toprule

         % \textbf{Models} & \textbf{Swahili} & \textbf{English} & \textbf{Chinese} & \textbf{Bengali} & \textbf{German} & \textbf{Spanish} & \textbf{French} & \textbf{Japanese} & \textbf{Russian} & \textbf{Thai} & \textbf{Avg.} \\ 
         \textbf{Lang.}  &  \textbf{Sw} & \textbf{Zh} & \textbf{Bn} & \textbf{De} & \textbf{Es} & \textbf{Fr} & \textbf{Ja} & \textbf{Ru} & \textbf{Th} \\ 
         \midrule
Agree. & 93.4 &
 94.3&
 93.8&
 92.7&
 93.8&
 96.9&
 92.4&
 93.2&
 92.9  \\

               \bottomrule
    \end{tabular}
    \end{adjustbox}
     \caption{Human agreement rate of each language in MSVAMP.}
     \label{nmsvamp_agreement}
     \vspace{-5pt}
\end{table}. The high agreement rates prove the reliable translation quality.
% \clearpage
% \section{Appendix}

\begin{table*}[!t]\footnotesize
\centering
\small

\begin{tabular}{l|p{0.8\linewidth}}
% \begin{tabular}{lc}
\toprule
\multirow{3}{*}{\textbf{ Input Prompts}} &   Below is an instruction that describes a task. \texttt{\string\n}
Write a response that appropriately completes the request in \{ \textit{language} \}. Please answer in \{ \textit{language} \}.\texttt{\string\n} \texttt{\string\n} \#\#\# Instruction: \texttt{\string\n} \{\textit{query}\}\texttt{\string\n}\texttt{\string\n} \#\#\# Response:   	\\

	% \midrule

\bottomrule
\end{tabular}

% English translation is only given for reading.} 
\caption{ Training and testing prompts in our experiments. }
\label{table:instruction}
% \vspace{-5mm}
\end{table*}

\section{Discussion}
\label{discussion}
\begin{table*}[t]
    \centering
    % \tiny
    
    % \small
    \begin{adjustbox}{width=1.0\linewidth}
    \begin{tabular}{l|cccccccccc|c}
    \toprule
  
         % \textbf{Models} & \textbf{Swahili} & \textbf{English} & \textbf{Chinese} & \textbf{Bengali} & \textbf{German} & \textbf{Spanish} & \textbf{French} & \textbf{Japanese} & \textbf{Russian} & \textbf{Thai} & \textbf{Avg.} \\ 
         \textbf{Models}  & \textbf{En} & \textbf{Sw} & \textbf{Zh} & \textbf{Bn} & \textbf{De} & \textbf{Es} & \textbf{Fr} & \textbf{JA} & \textbf{Ru} & \textbf{Th} & \textbf{Avg.} \\ 
         \midrule
%            \multicolumn{12}{c}{\textit{\textbf{Test On MGSM}}} \\

% \midrule[0.05pt]
 % \textbf{\ModelNameP}-LoRA &  30.4& 15.2&23.6& 10.4 & 22.8 & 24.8 & 26.4 & 18.0& 22.0& 14.8 & 20.8 \\
 LLaMA 2 (En) & 43.2& 5.2& 22.4& \textbf{3.2}& \textbf{37.2}& 32.4& 34.4& 15.2& 28.0& 4.8& 22.6 \\ 
\textbf{\ModelNameP} (En-Zh-Es) & 44.0& 3.6& \textbf{34.4}& 3.2& 33.6& \textbf{41.2}& \textbf{36.8}& \textbf{25.2}& \textbf{30.4}& 4.0& 25.6\\ 

   \textbf{\ModelNameP} (En-Sw-Th) &     \textbf{46.0}& \textbf{34.4}& 27.6& 2.4& 31.2& 35.2& 32.4& 22.4& 27.2& \textbf{36.8}& \textbf{29.6}\\ 
        \bottomrule
    \end{tabular}
    \end{adjustbox}
    \caption{Model Performances on MGSM test set. \textbf{\ModelNameP} (En-Zh-Es) refers to we only train \textbf{\ModelNameP} in three languages: English, Chinese and Spanish. Similarly,  \textbf{\ModelNameP} (En-Sw-Th) means the \textbf{\ModelNameP}  trained in English, Swahili and Thai.}
     \label{abl_lang}
\end{table*}

\subsection{RQ2: Targeted Language Training: Limited Broader Linguistic Reasoning Gains}

Our exploration into model training with select languages posits a notable inquiry: Can training with a subset of languages enhance mathematical reasoning across all languages? Engaging two high-resource languages, Spanish and Chinese, and two low-resource languages, Thai and Swahili, for mixed training sessions reveals pivotal insights. Seen in Table \ref{abl_lang}, while stark performance enhancements are witnessed in trained languages, notably in low-resource ones like Thai (surging from 4\% to 36.8\%), the model's efficacy varies in languages that are unseen in training: While there are improved outcomes in certain languages like Japanese and French, a corresponding decline is witnessed in others, such as German and Russian. This phenomenon might predominantly stem from the disparities in grammatical structures across different languages \cite{chen-etal-2023-structural}.

\subsection{RQ3: Training Variability: Beyond English-centric Questions}

Beyond the two training strategies explored in Section 3.3, we further probe alternative approaches to discern their influence on model performance. Hence, we examine two additional strategies: 1) A \textbf{Mix-Training} approach, where cross-training and parallel-training data are amalgamated for training; 2) An expansive \textbf{Mix-All} method that not only extends cross-training but also randomly pairs two languages from the \texttt{MGSM8KInstrucT}, thereby permitting questions and answers in the training data to traverse various linguistic combinations, effectively amplifying the original training data volume tenfold. The resulting models obtained by the above strategies are called \ModelName$^\mathcal{M}$ and \ModelName$^\mathcal{M}$-All, separately.

\begin{figure}[!t]
\centering
\includegraphics[width=0.95\linewidth]{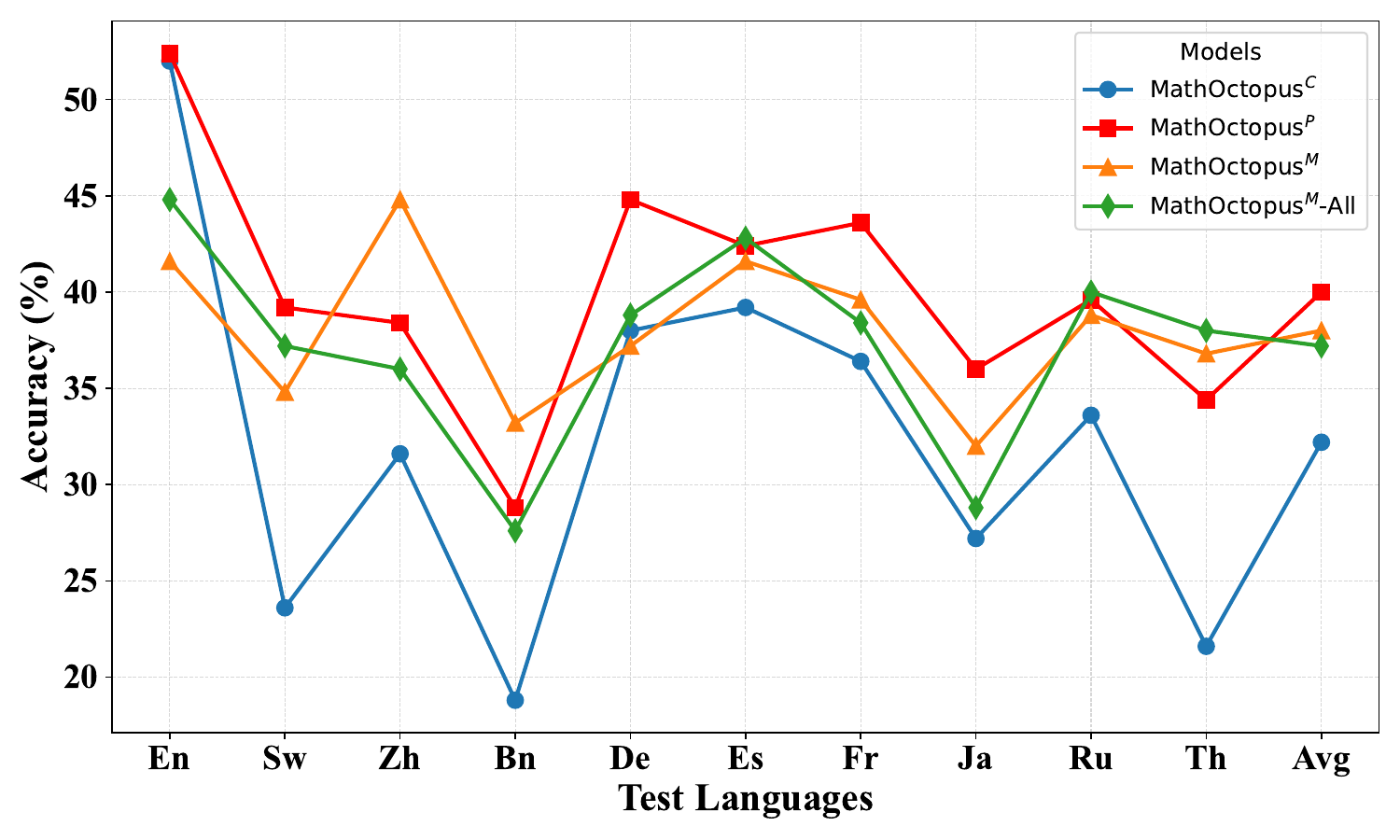}
% \vspace{-5pt}
\caption{Performances of  \ModelName-7B with different strategy on MGSM.
}
\label{streategy}
\vspace{-10pt}
\end{figure}

% \noindent %
% \begin{minipage}[c]{0.45\textwidth}
From the right-side Figure \ref{streategy}, it is evident that despite these two new training strategies respectively doubling and amplifying training volumes tenfold compared to original parallel-training and cross-training strategies, they do not surpass the results of \ModelNameP. Furthermore, the outcomes from \ModelName$^\mathcal{M}$-All slightly underperform \ModelName$^\mathcal{M}$. Such a phenomenon may arise because, although ``mix-all'' and ``mix-training''  expand the original data, the pre-existing data volume already suffices for the model to learn alignment and reasoning capabilities across different languages. An additional, rudimentary data expansion potentially induces overfitting, subsequently diminishing model performance.
% \end{minipage}%
%    \hfill

\subsection{Prompts for ChatGPT Translation}
\begin{table*}[!t]\footnotesize
\centering
\small

\begin{tabular}{p{0.95\linewidth}}
% \begin{tabular}{lc}
% \multirow{28}{*}{\textbf{Persona Prompts}} & 
\toprule

\textbf{Prompts}: You are a professional \{\texttt{lang}\} translator and spelling corrector. Please translate the given math question and its reasoning path into \{\texttt{lang}\}.\\
Below are examples: \\
Q: Weng earns \$12 an hour for babysitting. Yesterday, she just did 50 minutes of babysitting. How much did she earn? \\
P: Babysitting is \$12/hour = \$12/60 minutes = \$<<12/60=0.2>>0.2 per minute. Weng did babysitting for 50 minutes, so she earned \$0.2 x 50 = \$<<0.2*50=10>>10. \\

T-Q: Weng gana 12 dólares la hora por cuidar niños. Ayer cuidó niños durante 50 minutos. ¿Cuánto ganó? \\
T-P: Cuidar niños cuesta \$12/hora = \$12/60 minutos = \$<<12/60=0.2>>0.2 por minute. Weng cuidó niños durante 50 minutos, por lo que ganó \$0.2 x 50 = \$<<0.2*50=10>>10.
\\
Q: Julie is reading a 120-page book. Yesterday, she was able to read 12 pages and today, she read twice as many pages as yesterday. If she wants to read half of the remaining pages tomorrow, how many pages should she read? \\
P: Since today is the last day of the month, Julie would like to finish an entire book before tomorrow. She has read a total of 12 + 24 = <<12+24=36>>36 pages in two days. There are 120 - 36 = <<120-36=84>>84 pages left to be read. Hence, Julie should read 84/2 = <<84/2=42>>42 pages tomorrow. \\
T-Q: Julie está leyendo un libro de 120 páginas. Ayer pudo leer 12 páginas y hoy leyó el doble que ayer. Si quiere leer la mitad de las páginas restantes mañana, ¿cuántas páginas debería leer? \\
T-P: Como hoy es el último día del mes, a Julie le gustaría terminar un libro completo antes de mañana. Ha leído un total de 12 + 24 = <<12+24=36>>36 páginas en dos días. Quedan 120 - 36 = <<120-36=84>>84 páginas por leer. Por lo tanto , Julie debería leer 84/2 = <<84/2=42>>42 páginas mañana. \\
Please Keep in mind that: \\(1) keep the translations consistent for names of people and places within the sentences. \\ (2) Preserve the mathematical formula within the ``<< >>'' brackets when translating. \\ (3) You must translate the text into \{\texttt{lang}\}. \\(4) You must follow the output format with: "T-Q:... T-P:..."
\\
\bottomrule
\caption{Translation prompts in our experiments. } 
\label{table:transprompt}
\end{tabular}

\vspace{-5mm}
\end{table*}

\begin{table*}[!t]\footnotesize
\centering
\small

\begin{tabular}{p{0.95\linewidth}}
% \begin{tabular}{lc}
% \multirow{28}{*}{\textbf{Persona Prompts}} & 
\toprule

\textbf{Prompts}: Answer the following math probelm step by step in \{\texttt{lang}\}.\\
Below are examples: \\
Q: Julia played tag with 18 kids on Monday. She played tag with 10 kids on Tuesday. how many more kids did she play with on Monday than on Tuesday? \\
A: Let's think step by step. Julia playsed tag with 18 kids on Monday and 10 kids Tuesday, separately. So the amount of kids that she played with on Monday than on Tuesday is 18-10=8. So the answer is 8. \\
Q: Jack had 9 action figures and 10 books on a shelf in his room. later he added 7 more action figures to the shelf. how many more action figures than books were on his shelf ? \\
A: Let's think step by step. The amount of action figures that Jack had is 9+7=16. And Jack had 10 books. So the amount of action figures than books on his shelf is 16-10=6. So the answer is 7.
\\
\bottomrule
\caption{2-shot prompts of ChatGPT and GPT-4 in our experiments. } 
\label{table:testprompt}
\end{tabular}

\vspace{-5mm}
\end{table*}
Table \ref{table:transprompt} shows the translation prompt when we utilize ChatGPT to translate the data from English to other languages. When applying this prompt, we just need to replace the placeholder symbol \texttt{lang} with the specific language. Here, we use the Spanish translation examples in the prompt. 

\subsection{Generated Cases by \ModelName}

\begin{CJK*}{UTF8}{gbsn}
\begin{table*}[!t]\footnotesize
\centering
\small

\begin{tabular}{lp{0.85\linewidth}}
% \begin{tabular}{lc}
% \multirow{28}{*}{\textbf{Persona Prompts}} & 
\toprule
\multicolumn{2}{c}{Generated Cases in Chinese} \\
\midrule
\multirow{3}{*}{\textbf{Query}} & 珍妮特的鸭子每天下 16 颗蛋。她每天早上早餐时吃 3 颗，每天用 4 颗为自己的朋友做松饼。剩下的鸭蛋她每天拿去农贸市场卖，每颗新鲜鸭蛋卖 2 美元。她每天在农贸市场赚多少钱？\\
\midrule
\multirow{3}{*}{LLaMA 2} & 珍妮特每天吃早餐3颗鸭蛋，剩下的鸭蛋用于做松饼4颗，剩下的鸭蛋卖在农贸市场2颗，珍妮特每天在农贸市场赚2*2=<<2*2=4>>4美元。珍妮特每天卖 4 颗鸭蛋，所以她每天在农贸市场赚 4*4=<<4*4=16>>16 美元。{\color{red}\ding{55}}\\
\midrule
\multirow{3}{*}{\ModelName} & 珍妮特每天吃掉 3 + 4 = <<3+4=7>>7 颗鸭蛋。鸭子每天下 16 颗蛋，所以剩下 16 - 7 = <<16-7=9>>9 颗蛋。珍妮特每天卖掉 9 颗蛋，每颗蛋卖 2 美元，所以她每天赚 9 * 2 = <<9*2=18>>\textbf{18}。{\color{green}\ding{51}} \\
\midrule
\textbf{Answer} &\textbf{18} \\

\midrule
\multirow{2}{*}{\textbf{Query}} & 图卢兹的绵羊数量是查尔斯顿的两倍。查尔斯顿的绵羊数量是西雅图的 4 倍。如果西雅图有 20 只绵羊，图卢兹、查尔斯顿和西雅图一共有多少只绵羊？\\
\midrule
\multirow{3}{*}{LLaMA 2} & 查尔斯顿有 2 * 20 = <<2*20=40>>40 只绵羊。图卢兹有 40 * 2 = <<40*2=80>>80 只绵羊。西雅图和查尔斯顿和图卢兹一共有 20 + 40 + 80 = <<20+40+80=140>>140 只绵羊。{\color{red}\ding{55}}\\
\midrule
\multirow{2}{*}{\ModelName} & 查尔斯顿有 4 * 20 = <<4*20=80>>80 只绵羊。图卢兹有 2 * 80 = <<2*80=160>>160 只绵羊。总共，三个城市有 20 + 80 + 160 = <<20+80+160=260>>\textbf{260} 只绵羊。{\color{green}\ding{51}} \\
\midrule
\textbf{Answer} &\textbf{260} \\
\midrule
\midrule
\multicolumn{2}{c}{Generated Cases in Japanese} \\
\midrule
\multirow{2}{*}{\textbf{Query}} & リチャードは15階建てのマンションに住んでいる。各階には8部屋あり、建物の3/4が入居されている。建物で未入居の部屋は全部で何室あるか？\\
\midrule
\multirow{3}{*}{LLaMA 2} & 15階建てのマンションに8部屋ある各階は15*8=<<15*8=120>>120部屋になる。建物の3/4が入居されているので、入居されている部屋数は120*3/4=<<120*3/4=90>>90部屋になる。未入居の部屋は90-8=<<90-8=82>>82部屋である。 {\color{red}\ding{55}}\\
\midrule
\multirow{3}{*}{\ModelName} & 15階建てのマンションに8部屋あるため、リチャード住む15階建てマンションの部屋数は8*15=<<8*15=120>>120部屋建物の3/4が入居されているため、未入居の部屋は120*1/4=<<120*1/4=30>>30部屋。 {\color{green}\ding{51}} \\
\midrule
\textbf{Answer} &\textbf{30} \\

\midrule
\multirow{3}{*}{\textbf{Query}} & トゥルーズはチャールストンに比べ2倍の数の羊を飼っています。チャールストンはシアトルに比べ4倍の数の羊を飼っています。シアトルが飼っている羊の数が20頭だったとしたら、トゥルーズ、チャールストン、シアトルが飼っている羊の総数は何頭でしょうか？\\
\midrule
\multirow{5}{*}{LLaMA 2} & トゥルーズはチャールストンに比べ2倍の数の羊を飼っているので、トゥルーズが飼っている羊の数は2*4=<<2*4=8>>8頭です。チャールストンはシアトルに比べ4倍の数の羊を飼っているので、チャールストンが飼っている羊の数は4*20=<<4*20=80>>80頭です。シアトルが飼っている羊の数が20頭だったとしたら、トゥルーズ、チャールストン、シアトルが飼っている羊の総数は80+8+20=<<80+8+20=118>>118頭です。{\color{red}\ding{55}}\\
\midrule
\multirow{4}{*}{\ModelName} & シアトルが20頭の羊を飼っているなら、チャールストンは4*20 = <<4*20=80>>80頭の羊を飼っています。トゥルーズはチャールストンの2倍の数の羊を飼っているので、160頭の羊を飼っています。合計すると、トゥルーズ、チャールストン、シアトルは20+80+160 = <<20+80+160=260>>\textbf{260}頭の羊を飼っています。{\color{green}\ding{51}} \\
\midrule
\textbf{Answer} &\textbf{260} \\

\bottomrule
\end{tabular}
\caption{Generated responses compared between LLaMA and ours within 7B-level on MGSM. } 
\label{table:examples}

% \vspace{-5mm}
\end{table*}
\end{CJK*}

\begin{CJK*}{UTF8}{gbsn}

% \selectlanguage{russian}
\begin{table*}[!t]\footnotesize
\centering
\small

\begin{tabular}{lp{0.85\linewidth}}
% \begin{tabular}{lc}
% \multirow{28}{*}{\textbf{Persona Prompts}} & 
\toprule
\multicolumn{2}{c}{Generated Cases in German} \\
\midrule
\multirow{4}{*}{\textbf{Query}} & Toula ging in die Bäckerei und kaufte verschiedene Arten von Gebäck. Sie kaufte 3 Dutzend Donuts, die 68 US-Dollar pro Dutzend kosten, 2 Dutzend Mini-Cupcakes, die 80 US-Dollar pro Dutzend kosten und 6 Dutzend Mini-Cheesecakes für 55 US-Dollar pro Dutzend. Wie hoch waren die Gesamtkosten? \\
\midrule
\multirow{4}{*}{LLaMA 2} & 3 Dutzend Donuts kosteten 3 × 68 = <<3*68=204>>204 US-Dollar.2 Dutzend Mini-Cupcakes kosteten 2 × 80 = <<2*80=160>>160 US-Dollar.6 Dutzend Mini-Cheesecakes kosteten 6 × 55 = <<6*55=330>>330 US-Dollar.Die Gesamtkosten für alle drei Arten von Gebäck waren 204 + 160 + 330 = <<204+160+330=794>>794 US-Dollar. {\color{red}\ding{55}}\\
\midrule
\multirow{5}{*}{\ModelName} & Toula kaufte 3 Dutzend Donuts für 68 US-Dollar pro Dutzend, also kosten sie 3 * 68 = 204 US-Dollar.Toula kaufte 2 Dutzend Mini-Cupcakes für 80 US-Dollar pro Dutzend, also kosten sie 2 * 80 = 160 US-Dollar.Toula kaufte 6 Dutzend Mini-Cheesecakes für 55 US-Dollar pro Dutzend, also kosten sie 6 * 55 = 330 US-Dollar.Die Gesamtkosten betragen 204 + 160 + 330 = 694 US-Dollar. {\color{green}\ding{51}} \\
\midrule
\textbf{Answer} &\textbf{694} \\
% \midrule
% \midrule
% \multicolumn{2}{c}{Generated Cases in Russian} \\
% \midrule
% \multirow{2}{*}{\textbf{Query}} &В Тулузе в два раза больше овец, чем в Чарльстоне. В Чарльстоне в 4 раза больше овец, чем в Сиэтле. Сколько овец в Тулузе, Чарльстоне и Сиэтле вместе взятых, если в Сиэтле 20 овец?\\
% \midrule
% \multirow{3}{*}{LLaMA 2} & Чарльстон: 4(20)=80 овецТулуза: 2(80)=160 овецСиэтл: 20+160=<<20+160=180>>180 овецВсех городов вместе взятых: 180+160+20=<<180+160+20=460>>460 овец {\color{red}\ding{55}}\\
% \midrule
% \multirow{3}{*}{\ModelName} & В Чарльстоне 4*20 = <<4*20=80>>80 овец.В Тулузе 2*80 = <<2*80=160>>160 овец.Всего 160+80+20 = <<160+80+20=260>>260 овец. {\color{green}\ding{51}} \\
% \midrule
% \textbf{Answer} &\textbf{260} \\
\midrule
\midrule
\multicolumn{2}{c}{Generated Cases in French} \\
\midrule
\multirow{2}{*}{\textbf{Query}} & Jacques fait des séries de 3 sprints 3 fois par semaine. Il court 60 mètres à chaque sprint. Au total, combien de mètres court-il par semaine ? \\
\midrule
\multirow{2}{*}{LLaMA 2} & Jacques fait 3 * 3 = <<3*3=9>>9 séries de sprints par semaine.Par série, il court 9 * 60 = <<9*60=540>>540 mètres.Par semaine, il court donc 540 * 9 = <<540*9=4800>>4800 mètres. {\color{red}\ding{55}}\\
\midrule
\multirow{3}{*}{\ModelName} & Jacques fait des séries de 3 sprints 3 fois par semaine.Chaque série est de 3 sprints, donc il court 3*3=<<3*3=9>>9 sprints par semaine.Chaque sprint est de 60 mètres, donc il court 60*9=<<60*9=540>>540 mètres par semaine. {\color{green}\ding{51}} \\
\midrule
\textbf{Answer} &\textbf{540} \\
\midrule
\midrule
\multicolumn{2}{c}{Generated Cases in Spanish} \\
\midrule
\multirow{4}{*}{\textbf{Query}} & Los patos de Janet ponen 16 huevos por día. Ella come tres en el desayuno todas las mañanas y usa cuatro para hornear magdalenas para sus amigos todos los días. Vende lo que sobra en el mercado de productores diariamente a \$2 el huevo fresco de pato. ¿Cuánto gana en dólares todos los días en el mercado de productores?\\
\midrule
\multirow{7}{*}{LLaMA 2} & Janet comienza con 16 huevos frescos por día - 3 huevos frescos que come para desayuno = <<16-3=13>>13 huevos frescos que no se comen.Janet usa 13 huevos frescos - 4 huevos frescos para hornear magdalenas = <<13-4=9>>9 huevos frescos que no se comen.Janet vende 9 huevos frescos - 2 huevos frescos que se venden en el mercado de productores = <<9-2=7>>7 huevos frescos que se venden en el mercado de productores.Janet gana 7 huevos frescos * \$2/huevo fresco = \$<<7*2=14>>14 en el mercado de productores todos los días. {\color{red}\ding{55}}\\
\midrule
\multirow{4}{*}{\ModelName} & Janet come 3 huevos por día y usa 4 huevos por día, por lo que tiene 3 + 4 = <<3+4=7>>7 huevos que no usa para nada.Los patos de Janet ponen 16 huevos por día, por lo que tiene 16 - 7 = <<16-7=9>>9 huevos excedentes.Vende los 9 huevos excedentes por \$2 cada uno, por lo que gana 9 * 2 = \$<<9*2=18>>18 todos los días en el mercado de productores.{\color{green}\ding{51}} \\
\midrule
\textbf{Answer} &\textbf{18} \\

\bottomrule

\end{tabular}
\caption{Generated responses compared between LLaMA and ours within 7B-level on MGSM. } 
\label{table:examples2}
% \vspace{-5mm}
\end{table*}
\end{CJK*}

Table \ref{table:examples} and Table \ref{table:examples2} presens several generated examples in various languages by LLaMA and our models.
% \section{Example Appendix}
% \label{sec:appendix}

% This is an appendix.

\end{document}